# Synthesizing Customized Planners from Specifications

**Biplav Srivastava**                                                    BIPLAV@ASU.EDU
**Subbarao Kambhampati**                                                 RAO@ASU.EDU
*Department of Computer Science and Engineering*
*Arizona State University, Tempe, AZ 85287.*

## Abstract

Existing plan synthesis approaches in artificial intelligence fall into two categories –
domain independent and domain dependent. The domain independent approaches are ap-
plicable across a variety of domains, but may not be very efficient in any one given domain.
The domain dependent approaches need to be (re)designed for each domain separately, but
can be very efficient in the domain for which they are designed. One enticing alternative
to these approaches is to automatically synthesize domain independent planners given the
knowledge about the domain and the theory of planning. In this paper, we investigate
the feasibility of using existing automated software synthesis tools to support such synthe-
sis. Specifically, we describe an architecture called CLAY in which the Kestrel Interactive
Development System (KIDS) is used to derive a domain-customized planner through a
semi-automatic combination of a declarative theory of planning, and the declarative con-
trol knowledge specific to a given domain, to semi-automatically combine them to derive
domain-customized planners. We discuss what it means to write a declarative theory of
planning and control knowledge for KIDS, and illustrate our approach by generating a class
of domain-specific planners using state space refinements. Our experiments show that the
synthesized planners can outperform classical refinement planners (implemented as instan-
tiations of UCP, Kambhampati & Srivastava, 1995), using the same control knowledge. We
will contrast the costs and benefits of the synthesis approach with conventional methods
for customizing domain independent planners.

## 1. Introduction

Given the current state of the world, a set of desired goals, and a set of action templates,
"*planning*" involves synthesizing a sequence of actions which when executed from the initial
state will lead to a state of the world that satisfies all the goals (Fikes & Nilsson, 1990;
McAllester & Rosenblitt, 1991; Kambhampati, 1997b). Planning is known to be a combina-
torially hard problem, and a variety of approaches for plan synthesis have been developed
over the past twenty years. These approaches can be classified into two broad categories
– domain independent and domain dependent. Domain independent planners do not make
any assumptions about the planning domains, and can thus accept and solve planning prob-
lems from any domain. In contrast, domain specific planners are specifically designed for a
single domain and thus have the dynamics and control knowledge of the domain hard-coded.

The advantage of domain independent planning is that once a planning algorithm is
designed, it can be used in any domain by simply changing the action template that is
input to the algorithm. In contrast, domain-specific planners would have to be "modified"
or "re-designed" for each domain. On the flip-side, domain-specific planners tend to be more efficient in their designated domains than domain independent planners since the latter may not be able to effectively exploit the control knowledge of every domain.

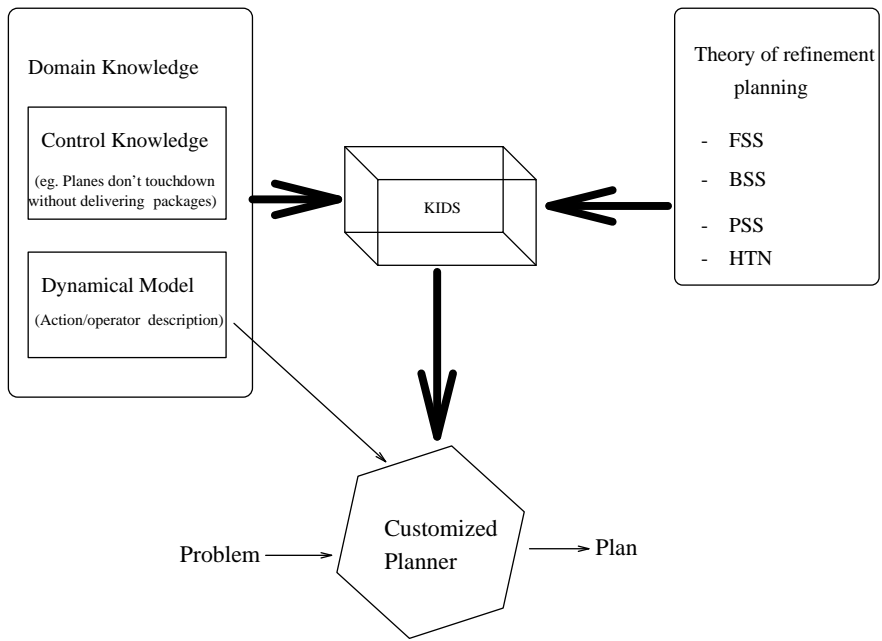

Figure 1: Architectural overview of planner synthesis with KIDS in the CLAY approach. The theories of refinement planning and domain knowledge are declaratively specified to KIDS which in turn combines them to produce a customized planner for the domain. The resulting planner, like conventional planners, can handle any planning problem from the domain. For a detailed description, see Section 3.

Not surprisingly, a significant amount of work in AI planning has been aimed at improving the performance of domain independent planners by dynamically customizing them to a given domain. This customization is done by providing the domain writer the ability to control the search of the planner, as is the case in task-reduction planning (Kambhampati & Srivastava, 1996; Kambhampati, 1995), or by using learning techniques (Kambhampati, Katukam, & Qu, 1996; Minton, 1990). Although several approaches have been developed for learning to improve planning performance, at present they are not an effective match for the efficiency of domain dependent planners.

One intriguing alternative is to automatically synthesize domain dependent planners given the knowledge about the domain and the theory of planning. In this paper, we investigate the feasibility of using existing automated software synthesis tools to support such a synthesis. We introduce the CLAY architecture which supports the synthesis of domain dependent planners using KIDS, a semi-automated software synthesis system. Specifically, as shown in Figure 1, a declarative theory of plan synthesis (theory of planning) is combined with the control knowledge specific to a given domain in a semi-automated software synthesis system called Kestrel Interactive Development System – KIDS (Smith, 1990, 1992a, 1992b) to derive a customized planner for the domain. We will draw the declarative theory of plan synthesis from domain independent planning techniques. Domain specific control knowledge will be expressed in terms of the types of plans that are preferred in the given domain.

Such an approach strikes a promising middle-ground between domain independent and domain dependent planners. The theories of planning are encoded independent of domains, and the domain control knowledge can be encoded independent of the specific planning theory being used. The customization step compiles the domain control knowledge into the planning algorithm and ensures that the resulting planners are able to exploit the structure of the domain.

## 1.1 Overview of the Synthesis Approach

As briefly mentioned above, the practicality of our approach is predicated on the availability of a software synthesis system capable of deriving code from formal specifications. KIDS is a powerful semi-automated system for development of correct and efficient programs from formal specifications. Given a domain theory and the input/output specification of a task, KIDS system helps in synthesizing a program capable of solving the task. Here, the term *theory* refers to any useful body of knowledge. *Task* refers to any assignment that is given to KIDS for solving and the solution is a *program* for that task. The input to KIDS is a task theory comprised of the task specification and a declarative description of useful concepts and rules to reason in the task space. In this research, we give planning as a task to KIDS and expect it to synthesize and return a planner as the solution. The planner can then take planning *problems* as input and return *results* (plans).

In order to support planner synthesis, we have to develop and input a theory of planning to KIDS. As discussed in (Kambhampati, 1997b), the traditional plan synthesis techniques can be described in terms of a common plan representation, with different planners corresponding to different ways of refining the partial plans such as progression, regression and plan-space refinements (see Section 2.2). Consequently, our planning theory will consist of a specification of the planning task (in terms of input and output data types) and one or more refinement theories. Since we are also interested in domain-customized planners, we have to provide the necessary domain knowledge to KIDS.

Given these inputs, KIDS semi-automatically synthesizes a program (in this case, a domain dependent refinement planner) using generic algorithm design tactics (such as branch and bound, global search). The resulting planner, like conventional planners, can handle any planning problem from the domain. See Section 3 for more details.

## 1.2 Outcomes

To understand the efficacy of plan synthesis in CLAY, in this paper, we concentrate on the synthesis of planners using state-space refinement theories.[1] Empirical evaluation shows that these synthesized planners can be very efficient. For example, in the blocks world domain where the goal was stack inversion, a KIDS synthesized planner solved a 14 blocks problem in under a minute. In the logistics domain, a problem with 12 packages, 4 planes and 8 places was solved in under a minute. Similarly, in the Tyre domain (Russell & Norvig, 1995), the "fixit" problem was solved in under a minute. To put these performance results in perspective, we compared KIDS' synthesized planners to a set of classical planners implemented as the instantiations the UCP planning system (Kambhampati & Srivastava, 1995). As described later, instantiations of UCP can emulate a spectrum of classical planners, including the popular SNLP planner (McAllester & Rosenblitt, 1991), by selecting the appropriate refinement. In our experiments, the best of the KIDS' synthesized planners outperformed the best of the UCP instantiations when given the same domain-specific information. We hypothesize that this is because KIDS can profitably fold-in the domain-specific control knowledge (i.e., the domain theory) into the planning code.

---

1. In future, we plan to extend our approach to plan-space and task-reduction refinements.

## 1.3 Organization

The rest of this paper describes the details of our approach, called the CLAY architecture for planner synthesis. The paper is organized as follows: after a brief review of traditional plan synthesis approaches and software synthesis on KIDS in Section 2, we walk through the CLAY framework in Section 3. Section 4 presents a discussion on the nature of planners synthesized by our approach. Section 5 empirically evaluates the synthesized planners and compares them to classical planners. Section 6 discusses related work. Section 7 describes our conclusions and discusses the costs and benefits of the synthesis approach.

## 2. Background

In this section, we briefly discuss relevant background on software synthesis with KIDS, and plan generation that will be needed to follow the rest of the paper.

### 2.1 Kestrel Interactive Development System

Before discussing KIDS, we start with some preliminaries on automated software synthesis. The holy grail of software synthesis is to :

- Produce highly reliable, adaptable software in a greatly reduced development time.

- Automate detail intensive tasks in software production that are largely non-creative in nature.

A program, or program segment, $P$, is correct with respect to an initial condition (assertion) $I$ and a final condition (assertion) $F$ if and only if whenever $I$ is true prior to the execution of $P$, and $P$ terminates, $F$ will be true after the execution of $P$ is complete. Using a formal specification of the task, a knowledge-base and an inference system, knowledge-based software synthesis proceeds with an iterative specification refinement process to specialize the general knowledge of program development (in the form of algorithm theories) to solve specific tasks on hand. The refinements are sound in that if the specification is correct, the synthesized program (code) will be correct.

KIDS is a program-transformation framework for the development of programs from formal specifications of a task. KIDS runs on Sun workstations and is built over REFINE, a commercial knowledge-based programming environment and a high-level language. The REFINE language supports first-order logic, set-theory, pattern matching and transformation rules. Refine provides a compiler that generates Common Lisp or C code for programs written in its logical specification language.

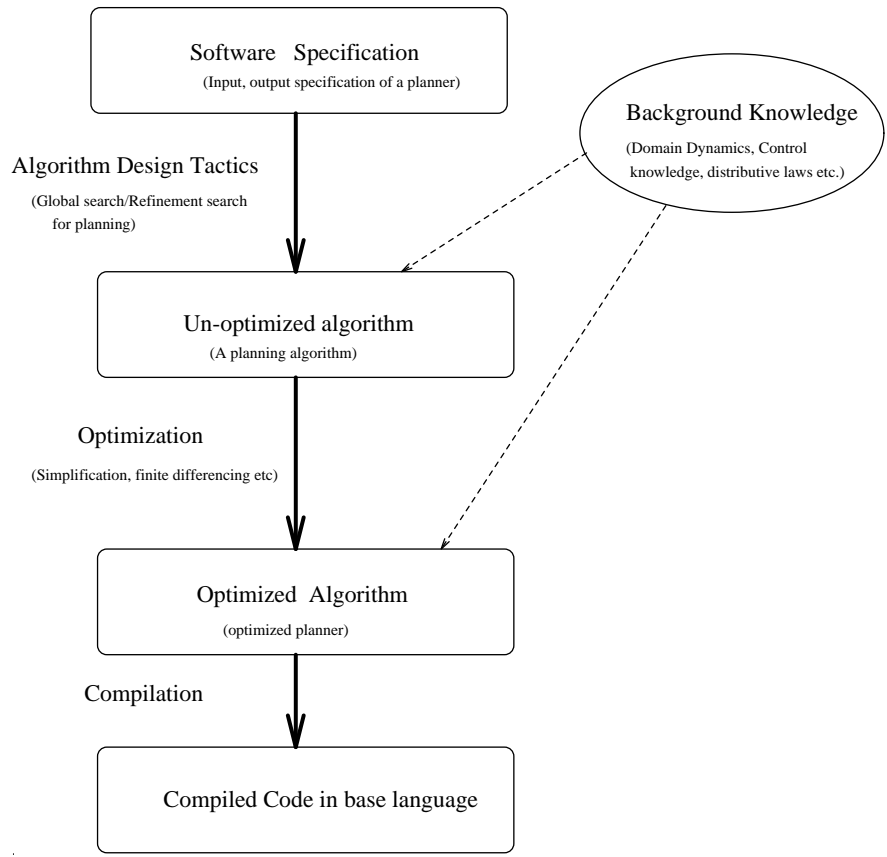

Figure 2: Overview of software synthesis process in KIDS

In the following, we describe the general steps involved in synthesizing software on KIDS. Figure 2 provides an overview of this process. The process is illustrated in more detail in Section 3 in the context of synthesis of customized planner code.

1. *Develop a task theory* to state and reason about the task. The user defines appropriate functions and types that describe the task and also gives laws that allow high-level reasoning about the defined functions. For planning, many planning theories (e.g., progression and regression) were written and relevant laws were specified. We also provided domain theories so that KIDS could perform specialized reasoning on planners it returned as solutions.

2. *Select and apply a design tactic* to select an algorithmic framework that should be used to implement the task specification. KIDS currently supports a variety of design tactics including problem reduction, divide and conquer, global search and local search. For planning, we use the global search design tactic because our formalization of classical planning is driven by refinement search which can be seen as a special case of global search (see below).

3. *Apply optimizations* to make the generated algorithm efficient. At first, the generated algorithm is well-structured and correct in that it can return all valid solutions, but it can be very inefficient. The algorithm is optimized through specification reduction techniques such as simplification, partial evaluation and finite-differencing.

4. *Compile* the algorithm to produce a program in the base language.

The domain theories and specifications are written in REFINE, and KIDS synthesizes and optimizes the algorithms in the same language. To transform specifications into programs as well as to optimize the programs, KIDS uses a form of deductive reasoning called *"directed inference"* to reason about the task specification and domain theory.

The KIDS system has been used to derive a variety of programs in the past. Of particular interest to us is the work on deriving efficient scheduling software (Smith & Parra, 1993; Burstein & Smith, 1996), as the success of these programs provided initial impetus for our own research.

## 2.2 Theories of Plan Synthesis

As mentioned earlier, using KIDS to derive planning software in CLAY involves figuring out (a) how declarative theories for different types of classical planning are specified and (b) what algorithmic design templates are best suited to planner synthesis. (Kambhampati, 1997b) provides an overview of traditional plan synthesis approaches. As discussed there, plan synthesis approaches come in many varieties with very little superficial commonality between them. In the last few years, we have developed a unifying framework that subsumes most of these approaches (Kambhampati & Srivastava, 1995; Kambhampati, Knoblock, & Yang, 1995; Kambhampati, 1997b). In this framework, plan synthesis is modeled as a process of searching in a space of sets of action sequences. These sets are represented compactly as collections of constraints called "partial plans." The search process first attempts to extract a result (an action sequence capable of solving the problem) from the partial plan, and when that fails, "refines" (or splits) the partial plan into a set of new partial plans (each corresponding to sets of action sequences that are subsets of the action sequence set corresponding to the original partial plan), and considers the new plans in turn. The existing domain independent plan-synthesis algorithms correspond to four different ways of refining partial plans. These are known, respectively, as Forward State Space or progression refinement (FSS), Backward State Space Refinement or regression refinement (BSS), Plan Space Refinement (PSS) and Task-Reduction Refinement. STRIPS (Fikes & Nilsson, 1990) is an example of a planner using the FSS refinement, TOPI (Barrett & Weld, 1994) uses the BSS refinement, SNLP (McAllester & Rosenblitt, 1991) uses the plan-space refinement and NONLIN (Tate, 1977) uses the task-reduction refinement. Given this background, the declarative theory of plan generation in CLAY corresponds to theories of the refinements. The algorithm tactic underlying plan generation corresponds to "refinement search." The KIDS system supports an algorithm tactic called "global search" (Smith, 1992a) which can be seen as a generalization of this refinement search.

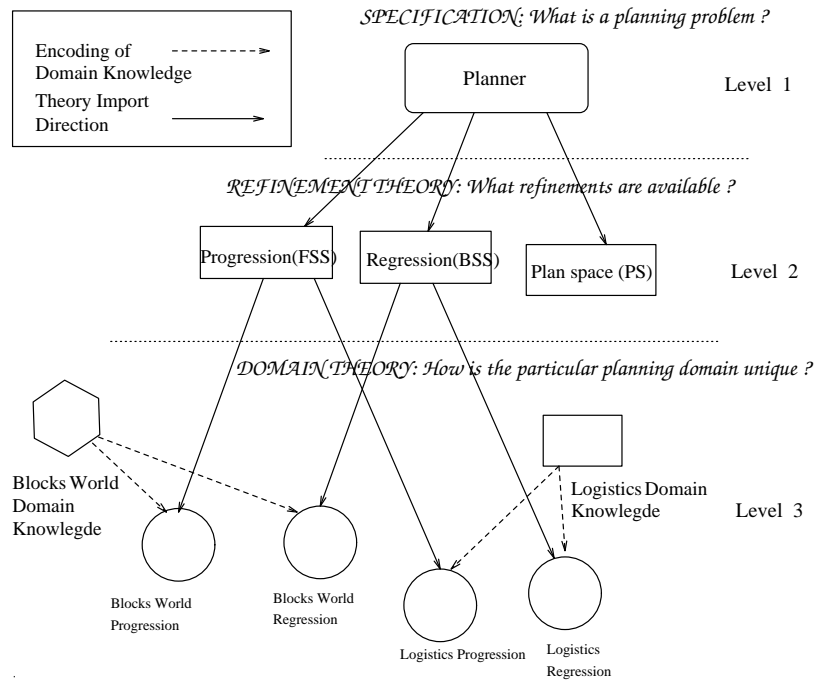

Figure 3: The CLAY architecture for writing planning theory. Each level answers a question relevant to that level of planning detail. CLAY uses KIDS' feature of theory import to modularize the domain-specific planning theory

## 3. Developing a Planner from Declarative Specification: The CLAY Architecture

Figure 1 summarizes how KIDS is used to synthesize a domain-specific refinement planner. The domain knowledge consists of a dynamical model and control knowledge. The dynamical model is specified in the form of actions (also called operators) that define legal transformations from one state of the world to another. Control knowledge is a set of domain-specific criteria that helps the planner decide if a plan $P_1$ is better than $P_2$ and is intended to make search more efficient. An example of control knowledge is that in a logistics domain where some packages have to be moved to their destinations using airplanes, planes should not touchdown at a location if they have no packages to pickup or deliver.

Refinement planning and domain control knowledge are brought together in the CLAY architecture for writing declarative domain-specific planning theory as summarized in Figure 3. To specify a planning task, a plan representation is selected and the constraints that should be satisfied by a solution plan are enumerated in the planner specification. The planner specification is dependent on the plan representation but is independent of the refinement needed for search. A refinement strategy uses the planner specification and defines how children nodes are generated from a given partial plan, what the goal test will be, and also explicates any refinement specific search pruning tests. The refinement and the specification together form the *planning theory*. To obtain a domain dependent planner, all one needs to do is import any planning theory and provide some relevant domain-specific planning control knowledge that provides a preference structure among partial plans and competing solutions. An interesting special case is when one specifies a *generic* domain knowledge to the effect that all the plans are equally good in the domain. In such a case, based on the refinement used, one gets a FSS, BSS, PSS or hybrid (if multiple refinements are used), general-purpose planner.

Each level in the directed tree in Figure 3 represents an abstraction of the planning task. At the root of the tree (Level 1), only a description of a planning task is required without specifying what refinements strategies should be used. At Level 2, the refinements are specified but no assumption is made about the domain. Next, characteristics of the domain are provided at Level 3. A progression (FSS) blocks world planner is different from a progression logistics domain planner only in terms of the domain knowledge. On the other hand, a progression blocks world planner is different from a regression (BSS) blocks-world planner only in terms of the refinement used.

As stated above, to ensure flexibility, the control knowledge should not change when different refinements are used and thereby represent substantial flexibility. But in practice, since control knowledge helps prune children nodes produced by a refinement, pruning may be more effective if the control knowledge is encoded depending on the refinement. There can also be a middle ground that we have not implemented: domain control knowledge may be represented in an intermediate form depending on the partial plan representation. Each refinement can provide, in addition to a termination test, a conversion function to transform the control knowledge into the refinement specific form.

## 3.1 Representing Domain Operators

We now discuss how the world state is represented and how the domain operators define state transformations. In classical planning, the world is modeled in terms of a set of "state variables." Each state of the world corresponds to a particular assignments of values to these variables. The actions are described in terms of the specific variable-value combinations that are needed for them to be applicable, and the variable-value combinations they will enforce after execution. Two variants of this general modeling approach have become popular in the planning community. The first, called the STRIPS representation (Fikes & Nilsson, 1990) represents the world in terms of ground atoms in a first order logic. The action applicability conditions and effects are also described in terms of conjunctions of ground atoms. The second variant (Backstrom & Nebel, 1993) models the world and actions directly in terms of multi-valued state variables and their values. Since STRIPS representation can be seen as a state-variable model with boolean state-variables, and since any multi-valued state-variable system can be converted into an equivalent boolean state-variable system, the two representations are equivalent in expressive power.

We chose the state-variable representation for our implementation since this can be directly mapped on to the primitive data structures supported by KIDS. Figure 4 shows the action of moving block A from block B to the top of block C in STRIPS and state-variable representation. A blocks world domain is an environment in which some blocks are placed on a table or on top of other blocks and the problems involve stacking them in

**STRIPS representation**

Action: $move(A, B, C)$
Prec: $clear(A) \wedge clear(C) \wedge \wedge on(A, B)$
Post: $on(A, C) \wedge clear(B) \wedge \neg clear(C)$
$\wedge \neg on(A, B)$

**Muti-valued state-variable representation**

Action: $move(A, B, C)$
Pre: $\langle B, True, 0, False, 0, True \rangle$
Post: $\langle C, True, 0, True, 0, False \rangle$

(Where a state is a 6-tuple
⟨ pos-A, clr-A, pos-B, clr-B, pos-C, clr-C ⟩,
and the zeroes represent "don't care" values)

Figure 4: Different representations of "Move A from B to C"

some desired configuration. For the purpose of exposition, we are showing values of state variables corresponding to block positions (e.g., pos-A) by symbols 'B', 'C', etc. and clear conditions (e.g., clr-A) by True or False. In practice, we map all the valid values of state variables to integers.

### 3.2 Specification of a Planner

KIDS uses a functional specification and programming language augmented with set-theoretic data types. A specification of the task (Smith, 1992a) is represented by a quadruple $F = \langle D, R, I, O \rangle$ where $D$ is the input type satisfying the input condition, $I : D \to boolean$. The output type is $R$ and the output condition, $O : D \times R \to boolean$, defines a feasible solution. If $O(x,z)$ holds, then $z$ is a feasible solution with respect to input $x$. The specification of a program follows the template:

> **function** $F$ (x :$D$) : $set(R)$
>
> **where** $I$(x)
>
> **returns** { z | $O$ (x,z) }
>
> = $Body(x)$

A specification for program $F$ is consistent if for all possible inputs satisfying the input condition, the body produces a feasible solution, i.e., $\forall(x : D)\exists(z : R)(I(x) \implies O(x, z))$.

Within this view, a planner takes as inputs an initial state, a goal state and an operator list. The operators are assumed to define state transitions from valid states to valid states. A specification for the planning task is: given the initial state, the goal state and the operator list, return a sequence of operators (plan) such that:

- TERMINATION-TEST: The goals must hold in the final state resulting from the execution of the plan. (We are only considering planning problems in which the goal is to make all state-variables achieve specified values, i.e., goals of achievement).

- DOMAIN-INDEPENDENT-PRUNING-TEST: The plan passes the domain independent pruning tests. Each planning refinemen can specify conditions under which a partial-plan cannot lead to a desirable solution; and any partial plan satisfying such a pruning

```
function PLANNER
  (INIT:  seq(integer), GOAL: seq(integer),
   OPERS: seq(tuple(seq(integer), seq(integer))))
  returns
    (PLAN: seq(integer)
       | range(PLAN) subset {1 .. size(OPERS)}
           & GOODNESS-TEST(VISITED-STATES
                              (PLAN, INIT, GOAL, OPERS),
                            INIT, GOAL)
           & NO-MOVES-BACK(VISITED-STATES
                              (PLAN, INIT, GOAL, OPERS),
                            INIT, GOAL)
           & GOAL-TEST(VISITED-STATES
                          (PLAN, INIT, GOAL, OPERS),
                        INIT, GOAL))
```

Figure 5: A declarative specification for planning

test can be eliminated from further consideration. For example, the state-loop pruning heuristic in forward state-space refinement says that if we have a partial plan such that the plan-state after executing operator $O_2$ is a subset of the state following an earlier operator $O_1$, then such a partial-plan can be pruned.

- DOMAIN-DEPENDENT-PRUNING-TEST: The plan passes the additional domain-specific pruning tests.

The last one is the hook through which domain specific control knowledge is introduced. In the current implementation of CLAY, we use domain knowledge for rejecting undesirable partial plans. In Section 7.1, we discuss ways in which our implementation can be extended to support other uses of domain control knowledge.

The specification of the planning task above is declarative in that it states what constraints must be satisfied in the resulting plan produced by a planner when given a planning problem. It does not suggest any algorithm that should be used to obtain the results. Algorithmic decisions will be made in the program development phase of KIDS.

An example of top-level specification of planning task (in REFINE) is shown in Figure 5. In this specification, the input condition, $I$, is true, the input data type, $D$ includes INIT, GOAL and OPERS and the output condition $R$ is PLAN and the output condition $O$ consists of GOODNESS-TEST, GOAL-TEST, NO-MOVES-BACK. NO-MOVES-BACK is a domain independent pruning tests whereas GOODNESS-TEST is a domain dependent pruning test.

Delving deeper into representation detail, we represent a plan as a sequence of indices in the operator list (i.e., sequence of operator identifiers). For state-space planning, the *state sequence* corresponding to the partial-plan is produced by function VISITED-STATES and the goal test, domain independent pruning tests and domain dependent pruning tests are done on the state-sequence by functions GOAL-TEST, NO-MOVES-BACK and GOODNESS-TEST respectively. The state-variables take integer values. Consequently, our initial and goal states are a sequence of integers.

In words, the specification in Figure 5 says that a partial plan is a sequence of integral indices (of operators) and so the indices must not be more than the size of operator list. Valid plan is one whose corresponding state sequence (produced by VISITED-STATES) satisfies the GOAL-TEST, NO-MOVES-BACK and GOODNESS-TEST[2].

In the context of forward-state space refinement (FSS), VISITED-STATES returns the states obtained by the successive application of the operators in the partial plan to the initial state and the resulting states thereafter. GOAL-TEST signals that the goal has been achieved; for FSS refinement it involves checking that the last state in the state-sequence is the goal state. The NO-MOVES-BACK function tests state looping; forward state-space looping checks if the state after executing operator $O_j$ (STATE $S_j$) is a subset of the state following an earlier operator $O_i(i < j)$ (STATE $S_i$), this partial-plan can be pruned[3].

The function GOODNESS-TEST checks for possible redundancy in the state sequence corresponding to the current partial plan based on domain characteristics. Let us explain it in the context of the blocks world domain. We can specify any reasonable checks for the blocks world as long as they do not make the planner lose a desired solution. Below, we present two GOODNESS-TESTs:

- (Heuristic H1: LIMIT USELESS MOVES) If a block moves between states i and (i+1), it must not change position between states (i+1) and (i+2). The motivation behind this check is to prevent blocks from being moved around randomly in successive moves.

- (Heuristic H2: MOVE VIA TABLE) A block can only move from its initial state to the table and from table only to its goal position. This is motivated by the fact that a polynomial time approximate algorithm for solving blocks world planning problems involves putting all blocks on table first, and then constructing each of the goal configuration stacks bottom-up.[4]

### 3.3 Implementing the Specification using Global Search

As discussed in Section 2, we need to select an algorithm design tactic to implement the task specification in KIDS. One of the design tactics provided by KIDS is *global search*. The basic idea of global search is to represent and manipulate sets of candidate solutions. The principal operations are to extract candidate solutions from a set and to split a set into subsets. Derived operations include various filters which are used to eliminate sets containing no feasible or optimal solutions. Global search algorithms work as follows: starting from an initial set of potential solutions, that contains all desired solutions to the given problem instance, the algorithm repeatedly extracts solutions, splits sets and eliminates sets via filters until a candidate solution can be drawn from one of the sets. Sets of solutions are represented implicitly by data structures called descriptors, and splitting is done by adding mutually exclusive and exhaustive sets of constraints to the descriptors. The process can be described as a tree search in which a node represents a set of candidates and an arc represents the split relationship between a set and its subset. For complete details, readers are referred to (Smith, 1992a).

---

2. REFINE code of all the referenced functions is shown in Appendix A.

3. Actually, we test a slightly more general condition that if the state after executing operator $O_j$ (STATE $S_j$) is *weaker* than a state following an earlier operator $O_i$ (STATE $S_i$), then this partial-plan can be pruned. $S_j$ is weaker than $S_i$ if every state-variable with assigned value in state $S_i$ has that same assigned value in state $S_j$. By specifying weakness rather than subset as the relationship between states to decide domain independent pruning in state space planning, we allow the synthesized planner to deal with a partially specified initial state. The planner will work correctly as long as all the state-variables that are required for reasoning are specified in the initial state.

4. Our pruning test alone doesn't guarantee polynomial algorithm since the order in which the blocks are to be put on the table or later at the goal positions is not specified in the pruning heuristic.

The KIDS' global search paradigm is a general form of the refinement search model used to unify classical planners in UCP (Kambhampati & Srivastava, 1995). Specifically, the partial plans correspond to descriptors and the refinements correspond to splitting strategies. To use global search to implement the planner specification, we need to select a suitable representation for sets of potential solutions (which, in the refinement view of planning, are essentially the partial plans). The global search tactic would then set up a search algorithm that can split a solution-set and extract solutions that meet the problem specification. KIDS provides global search tactics for primitive data-types such as sequences, sets and maps. If a complex data type is needed to represent the potential solution set of a task, the user must implement a global search tactic for it.

Since we are interested in state-space planners initially, we chose to represent the partial plan as a sequence of operators (actually sequence of operator indices). This allowed us to use KIDS global search theory for finite sequences.

## 3.4 Specifying Distributive and Monotonic laws

One aspect of KIDS specification that is slightly unintuitive to new users is the need to specify distributive and monotonic laws on all the operations used in the input/output specification (e.g., NO-MOVES-BACK in the specification of the progression planner shown in Figure 5). Distributive laws state how a specific operation distributes over other operations (e.g., $(A + B) \times C \equiv (A \times C) + (B \times C)$), while monotonic laws provide a set of boundary conditions (e.g., $A + 0 = A$). Such laws should be explicitly stated for all operations involved in the specification to support instantiation of design tactics, as well as optimization of generated code. Specifically, KIDS has a directed-inference engine called RAINBOW which uses the task specification and the distributed laws specified by the user to simplify and reformulate the expressions in the synthesized code. Deductive inference is the primary means by which KIDS reasons about the task specification in order to apply design tactics, and optimize the code, and derive necessary pruning conditions. Distributive and monotonic laws indirectly provide KIDS with information on alternative ways of defining predicates.

A useful heuristic in writing laws is that they should be simple, normally expressed in terms of the main function and perhaps another function to handle boundary cases (called cross-functions; see below). As an example, consider Figure 5 where function NO-MOVES-BACK is a domain independent pruning test used in the specification of progression planners. Recall that NO-MOVES-BACK is called on a state sequence and checks that a later state is not a subset of (or *weaker* than) an earlier state. Calling NO-MOVES-BACK on a sequence $S$ which is a concatenation of state sequences $S_1$ and $S_2$ is equivalent to calling it on $S_1$ and $S_2$ and testing that no state in $S_1$ is weaker than a state in $S_2$. Notice that the first two tests can be handled by NO-MOVES-BACK itself but the last test needs a new function. We call this new function a cross-function for NO-MOVES-BACK (CROSS-NO-MOVES-BACK). The monotonic laws for NO-MOVES-BACK include:

- NO-MOVES-BACK over a singleton state sequence is true.

- NO-MOVES-BACK over an empty state sequence is false (useful when the selected action is not applicable).

Similarly, the distributive laws for NO-MOVES-BACK include:

- NO-MOVES-BACK over sequence concatenation is equal to NO-MOVES-BACK on $S_1$, NO-MOVES-BACK on $S_2$ and CROSS-NO-MOVES-BACK on $S_1$ and $S_2$.

- When a state $A$ is prepended to a state sequence $S$, applying NO-MOVES-BACK over such a combined state sequence is equivalent to applying CROSS-NO-MOVES-BACK on the singleton sequence $[A]$ and $S$, and NO-MOVES-BACK on $S$.

- When a state $A$ is appended to a state sequence $S$, NO-MOVES-BACK over such a combined state sequence is the same as NO-MOVES-BACK on $S$ and CROSS-NO-MOVES-BACK on $S$ and the singleton sequence $[A]$.

All these rules, while reasonably obvious to us, are nonetheless very crucial for the effectiveness of KIDS as they help it in reformulating and optimizing the generated code. An example of their use, as we will see in Section 3.5.1, occurs in Figure 7 where an "if [condition] - [then] - [else]" statement gets simplified to just the "[then]" part because all the conditions in the "[condition]" can be proved to be true in the context of the input specification, given the distributive laws.

Using the task specification, the selected design tactic – global search, and distributive laws, KIDS produces a correct but naive code, as shown in Figure 6. The code is naive because the same checks (for example CROSS-NO-MOVES-BACK in Figure 6) are computed repeatedly even if they are true from their context. We discuss some methods to optimize the code in the next section.

## 3.5 Program Optimization

This section explains how the initial planner code, generated by KIDS, is optimized. Readers not familiar with automated software synthesis literature might want to skip this section on first read, and revisit it later for more details.

The first code produced by KIDS (shown in Figure 6) is well-structured but very inefficient. There are several opportunities for optimization and KIDS provides tools for program optimization. The code can be compiled and executed at any stage of optimization. Now, we briefly summarize the program optimizations used to achieve efficiency.

### 3.5.1 CONTEXT INDEPENDENT SIMPLIFIER

This method simplifies an expression independent of its surrounding context. There are two possibilities for context independent simplification:

- In the first case, a set of equations are treated as left-to-right rewrite rules that are fired exhaustively until none apply. Distributive laws are also treated as rewrite rules. An example application of a rewrite rule is:

    *(if true then P else Q)* $\implies P$

```
function PLANNER
 (INIT: seq(integer), GOAL: seq(integer),
  OPERS: seq(tuple(seq(integer), seq(integer))))
 returns
  (PLAN: seq(integer)
  | range(PLAN) subset {1 .. size(OPERS)}
    & GOODNESS-TEST
        (VISITED-STATES(PLAN, INIT, GOAL, OPERS),
         INIT, GOAL)
    & NO-MOVES-BACK
        (VISITED-STATES(PLAN, INIT, GOAL, OPERS),
         INIT, GOAL)
    & GOAL-TEST
        (VISITED-STATES(PLAN, INIT, GOAL, OPERS),
         INIT, GOAL))
 = if NO-MOVES-BACK
        (VISITED-STATES([], INIT, GOAL, OPERS),
         INIT, GOAL)
    & GOODNESS-TEST
        (VISITED-STATES([], INIT, GOAL, OPERS),
         INIT, GOAL)
   then PLANNER-AUX(INIT, GOAL, OPERS, [])
   else undefined
```

```
function PLANNER-AUX
 (INIT: seq(integer), GOAL: seq(integer),
  OPERS: seq(tuple(seq(integer), seq(integer))),
  V: seq(integer)
  | range(V) subset {1 .. size(OPERS)}
    & GOODNESS-TEST
        (VISITED-STATES(V, INIT, GOAL, OPERS),
         INIT, GOAL)
    & NO-MOVES-BACK
        (VISITED-STATES(V, INIT, GOAL, OPERS),
         INIT, GOAL))
 : seq(integer)
 = if ex (PLAN: seq(integer))
     (GOAL-TEST
        (VISITED-STATES(PLAN, INIT, GOAL, OPERS),
         INIT, GOAL)
     & NO-MOVES-BACK
        (VISITED-STATES(PLAN, INIT, GOAL, OPERS),
         INIT, GOAL)
     & GOODNESS-TEST
        (VISITED-STATES(PLAN, INIT, GOAL, OPERS),
         INIT, GOAL)
     & range(PLAN) subset {1 .. size(OPERS)}
     & PLAN = V)
   then some (PLAN-1: seq(integer))
     (GOAL-TEST
        (VISITED-STATES(PLAN-1, INIT, GOAL, OPERS),
         INIT, GOAL)
     & NO-MOVES-BACK
        (VISITED-STATES(PLAN-1, INIT, GOAL, OPERS),
         INIT, GOAL)
     & GOODNESS-TEST
        (VISITED-STATES(PLAN-1, INIT, GOAL, OPERS),
         INIT, GOAL)
     & range(PLAN-1) subset {1 .. size(OPERS)}
     & PLAN-1 = V)
   else some (PLAN-2: seq(integer))
     ex (NEW-V: seq(integer))
      (PLAN-2 =
         PLANNER-AUX(INIT, GOAL, OPERS, NEW-V)
      & DEFINED?(PLAN-2)
      & GOODNESS-TEST
          (VISITED-STATES(NEW-V, INIT, GOAL, OPERS),
           INIT, GOAL)
      & NO-MOVES-BACK
          (VISITED-STATES(NEW-V, INIT, GOAL, OPERS),
           INIT, GOAL)
      & ex (I: integer)
          (NEW-V = append(V, I)
           & I in {1 .. size(OPERS)}))
```

Figure 6: First progression blocks world planner code synthesized by KIDS. Notice that the code is inefficient because the same checks (for example, CROSS-NO-MOVES-BACK) are computed repeatedly, even if they are true from their context.

```
function PLANNER                              function PLANNER
  (INIT: seq(integer), GOAL: seq(integer),     (INIT: seq(integer), GOAL: seq(integer),
   OPERS: seq(tuple(seq(integer), seq(integer))))   OPERS: seq(tuple(seq(integer), seq(integer))))
  returns                                       returns
    (PLAN: seq(integer) |                         (PLAN: seq(integer) |
    range(PLAN) subset {1 .. size(OPERS)}        range(PLAN) subset {1 .. size(OPERS)}
    and  ...)                                    and ...)
 = if NO-MOVES-BACK                           = PLANNER-AUX
          (VISITED-STATES([], INIT, GOAL, OPERS),        (INIT, GOAL, OPERS, [], [INIT], INIT,
          INIT, GOAL)                                    size(OPERS))
       and GOODNESS-TEST
          (VISITED-STATES([], INIT, GOAL, OPERS),
          INIT, GOAL)
    then PLANNER-AUX
        (INIT, GOAL, OPERS, [], [INIT], INIT,
          size(OPERS))
    else undefined
```

Figure 7: Example of context independent simplification in KIDS. The code on the right size is the simplified version of that on the left.

```
function PLANNER-AUX                           function PLANNER-AUX
  (INIT: seq(integer), GOAL: seq(integer),      (INIT: seq(integer), GOAL: seq(integer),
   OPERS: seq(tuple(seq(integer), seq(integer))),  OPERS: seq(tuple(seq(integer), seq(integer))),
   V: seq(integer)                               V: seq(integer)
   | range(V) subset {1 .. size(OPERS)}          | range(V) subset {1 .. size(OPERS)}
   and GOODNESS-TEST(VISITED-STATES             and GOODNESS-TEST(VISITED-STATES
                  (V, INIT, GOAL, OPERS),                      (V, INIT, GOAL, OPERS),
                  INIT, GOAL)                                  INIT, GOAL)
   and NO-MOVES-BACK(VISITED-STATES             and NO-MOVES-BACK(VISITED-STATES
                  (V, INIT, GOAL, OPERS),                      (V, INIT, GOAL, OPERS),
                  INIT, GOAL))                                 INIT, GOAL))
  : seq(integer)                               : seq(integer)
 = if GOAL-TEST(                              = if GOAL-TEST(
          VISITED-STATES(V, INIT, GOAL, OPERS),         VISITED-STATES(V, INIT, GOAL, OPERS),
          INIT, GOAL)                                   INIT, GOAL)
       and NO-MOVES-BACK(VISITED-STATES            then (if ...) ...
                     (V, INIT, GOAL, OPERS),
                     INIT, GOAL)
       and GOODNESS-TEST(VISITED-STATES
                     (V, INIT, GOAL, OPERS),
                     INIT, GOAL)
       and range(V) subset {1 .. size(OPERS)}
    then (if ...) ...
```

Figure 8: Example of context dependent simplification in KIDS. The code on the right size is the simplified version of that on the left.

- In the second case, all occurrences of a local variable which is defined by an equality is replaced by the equivalent value:

$$\{C(x) \mid x = e \land P(x)\} \Longrightarrow \{C(e) \mid P(e)\}$$

Figure 7 shows an example of context-independent simplification where all the conditions of the if-condition are true from their respective distributive laws. Hence, the if-then-else statement is replaced by the then-part.

```
function PLANNER-AUX                          function PLANNER-AUX
  (INIT: seq(integer), GOAL: seq(integer),      (INIT: seq(integer), GOAL: seq(integer),
   OPERS: seq(tuple(seq(integer), seq(integer))), OPERS: seq(tuple(seq(integer), seq(integer))),
   V: seq(integer), VS: seq(seq(integer)),       V: seq(integer), VS: seq(seq(integer)),
   L-VS: seq(integer) |                           L-VS: seq(integer), NUM-OPS: integer
     SEQEQUAL(L-VS, last(VS))                         |
     and range(V) subset {1 .. size(OPERS)}        NUM-OPS = size(OPERS)
     and ...)                                       and ...)
= if GOAL-TEST(VS, INIT, GOAL) then V          = if GOAL-TEST(VS, INIT, GOAL) then V
  else some (PLAN-2: seq(integer))                else some (PLAN-2: seq(integer))
    exists(I: integer)                              exists(I: integer)
      (I in {1 .. size(OPERS)}                        (I in {1 .. NUM-OPS}
      and ...                                         and ...
```

Figure 9: Example of finite differencing (on size(OPERS)). The code on the right size is the result of finite differencing on that on the left.

```
 0. Focus Initialize PLANNER
 1. Tactic Global Search on PLANNER
 2. Simplify, context-independent-fast: if ## then ##
        else some (PLAN-2: ##...
 3. Simplify, context-dependent, forward-0, backward-4:
        ##(...) & ##(....
 4. Simplify, context-dependent, forward-0, backward-4:
        if ## else und...
 5. FD (general-purpose) VS =
        VISITED-STATES(V, INIT, GOAL, OPERS)
 6. FD (general-purpose) L-VS = last(VS)
 7. FD (general-purpose) NUM-OPS = size(OPERS)
 8. Abstract NEXT-STATE(L-VS, I, OPERS) into NS in
        ex (I: integer) (## in ## &...
 9. Simplify, context-independent-fast:
        if ## & ## then PLANNER-AUX(##, ##,...
10. Refine compile into Lisp: PLANNER-AUX, PLANNER
```

Figure 10: Derivation steps to generate a progression planner for blocks world domain.

### 3.5.2 Context Dependent Simplifier

This method is designed to simplify a given expression with respect to its surrounding context and is, thus, more powerful than context independent simplification. All the predicates that hold in the context of the expression are gathered by walking up the abstract syntax tree. The expression is then simplified with respect to the set of assumptions that hold in the context. In Figure 8, the function calls for NO-MOVES-BACK, GOODNESS-TEST and the range test in the "if" expression are redundant because their results follow from the input invariant (conditions listed after "|" and before ":") of the PLANNER-AUX function. So they are removed by this simplification.

```
function PLANNER
 (INIT: seq(integer), GOAL: seq(integer),
  OPERS: seq(tuple(seq(integer), seq(integer))))
 returns
  (PLAN: seq(integer)
  | range(PLAN) subset {1 .. size(OPERS)}
    & GOODNESS-TEST
       (VISITED-STATES(PLAN, INIT, GOAL, OPERS),
        INIT, GOAL)
    & NO-MOVES-BACK
       (VISITED-STATES(PLAN, INIT, GOAL, OPERS),
        INIT, GOAL)
    & GOAL-TEST
       (VISITED-STATES(PLAN, INIT, GOAL, OPERS),
        INIT, GOAL))
 = PLANNER-AUX
    (INIT, GOAL, OPERS, [], [INIT],
     size(OPERS), INIT)
```

```
function PLANNER-AUX
 (INIT: seq(integer), GOAL: seq(integer),
  OPERS: seq(tuple(seq(integer), seq(integer))),
  V: seq(integer), VS: seq(seq(integer)),
  NUM-OPS: integer, L-VS: seq(integer)
 | SEQEQUAL(L-VS, last(VS))
    & SEQEQUAL(VS,
       VISITED-STATES(V, INIT, GOAL, OPERS))
    & NO-MOVES-BACK
       (VISITED-STATES(V, INIT, GOAL, OPERS),
        INIT, GOAL)
    & GOODNESS-TEST
       (VISITED-STATES(V, INIT, GOAL, OPERS),
        INIT, GOAL)
    & range(V) subset {1 .. size(OPERS)}
    & NUM-OPS = size(OPERS))
 : seq(integer)
 = if GOAL-TEST(VS, INIT, GOAL) then V
    else some (PLAN-2: seq(integer))
      ex (NS: seq(integer), I: integer)
        (NS = NEXT-STATE(L-VS, I, OPERS)
         & CROSS-NO-MOVES-BACK(VS, [NS], INIT, GOAL)
         & CROSS-GOODNESS-TEST(VS, [NS], INIT, GOAL)
         & DEFINED?(PLAN-2)
         & PLAN-2
            = PLANNER-AUX
               (INIT, GOAL, OPERS, append(V, I),
                append(VS, NS), NUM-OPS, NS)
            & I in {1 .. NUM-OPS})
```

Figure 11: Final progression blocks world planner code synthesized by KIDS

### 3.5.3 FINITE DIFFERENCING

The idea behind finite differencing is to perform computations incrementally rather than repeat them from scratch every time. Let us assume that inside a function $f(x)$ there is an expression $g(x)$ and that $x$ changes in a regular way. In this case, it might be useful to create a new variable, equal to $g(x)$, whose value is maintained across iterations and which allows for incremental computation of $g(x)$ with the next $x$ value. Finite differencing can be decomposed into two more basic operations: abstraction and simplification (Smith, 1992a)

- First, the function $f$ is abstracted with respect to expression $g(x)$ adding a new parameter $c$ to parameter list of $f$ (now $f(x,c)$) and adding $c = g(x)$ as a new input invariant to $f$. All calls to $f$, whether recursive calls within $f$ or external calls, must now be changed to match the definition of $f$ i.e, $f(x)$ is changed to $f(x, g(x))$. In this process, all occurrences of $g(x)$ are replaced by $c$.

- If distributive laws apply to $g(h(x))$ yielding an expression of the form $h'(g(x))$ and so $h'(c)$, the new value of $g(h(x))$ can be computed in terms of the old value of $g(x)$ and this the real benefit in optimization.

Let us illustrate this process with an example. Suppose that some function $f(x)$ has a call for function $g(x)$ which returns the square of numbers and that variable $x$ varies linearly. Now, suppose that we are given a distributive law such as $g(x + 1) = g(x) + 2 * x + 1$. So, after finite-differencing, $f$ becomes $f(x,c)$ and the $g(x)$ call is replaced by $c + 2 * x + 1$.

An additional invariant $c = x * x$ will also be maintained for $f$. This new expression is computationally much cheaper than the original expression.

In Figure 9, finite differencing is performed on $size(OPERS)$. The argument of function PLANNER is expanded with the inclusion of NUM-OPS, the name entered by the user for the value of $size(OPERS)$. Only abstraction is done here as the number of operators available to the planner does not change during planning. Note that NUM-OPS represents the number of operators in a planning problem and this is a meaningful concept for planning. All instances of $size(OPERS)$ are replaced by NUM-OPS.

### 3.5.4 PROGRAM DERIVATION

Figure 10 shows a summary of the sequence of derivation steps carried out to obtain a blocks world domain-specific forward-state space planner. The final version of the planner code is shown in Figure 11.

In step 0, the top-level planner specification is selected and in step 1, selected design tactic is applied. Step 2 involves a context independent simplification, and steps 3 and 4 involve context dependent simplifications. Steps 5 through 8 cover finite differencing. Finally, an efficient planner code is compiled in step 10.

## 4. Discussion on Synthesized Planners

Section 3.5 used the synthesis of a progression planner for blocks world domain as a case study to walk-through the planner synthesis process leading to the final planner given in Figure 11. In our research, we have also considered regression (backward state space) planners. All the planners we have synthesized to date are summarized in Table 1. Although each of these planners differ in terms of the refinements they use, and the domains to which they are customized, Figure 12 attempts a pseudo-code description of a generic template to facilitate discussion of the synthesized planners as a group.

The main function PLANNER takes INITIAL_STATE, GOAL_STATE and OPERATOR_SET as inputs and in turn calls the recursive function PLANNER_AUX with all the inputs and an initial plan. All the pruning tests comprising the OUTPUT CONDITIONs are maintained as "invariants" (in that they must hold not only of the final plan, but also of the every partial plan leading to the final plan). The goal test, of course, need only hold for the final plan, and is thus not maintained as an invariant. Finite differencing leads to more invariants. Inside PLANNER_AUX, if a partial plan satisfies the goal test, it is returned. Otherwise, the partial plan is refined, invariants are incrementally tested in the new partial plan and PLANNER_AUX is called recursively.

First thing we note is that the pseudo-code template describes a planner for any planning domain employing any state-space refinement. Even the requirement that the synthesized planner be state-space is dictated by how the new partial plan (PARTIAL-PLAN2) is obtained (in this case by appending an operator to an operator sequence). It can be generalized to support other refinements, by modifying the operation (in the current case "append") that is used to build the new partial plan from the old one (with corresponding changes to the distributive laws to account for the new operation).

Second, we observe that all invariants are incrementally evaluated (see PLANNER_AUX). For example, to see if the the plan in $i^{th}$ iteration satisfies the NO-MOVES-BACK test, we only check if the latest state is duplicated by any of the previous states. We hypothesize that such incremental evaluation is the primary reason for the synthesized planner's efficiency. Once refinement and planning domain knowledge is available, context-dependent simplification may show that many of the tests made in the separate theories are in fact redundant. Moreover, incremental evaluations may be cheaper than complete evaluations of invariants if they are amenable to the operations of distribution and monotonicity over the abstract data-types.

Although domain independent planners can be given the same control knowledge that we give to KIDS during planner synthesis, our approach is expected to be superior in two ways:

1. Our approach separates the control knowledge acquisition from the specifics of the planner to some extent and this makes the acquisition process easier. In contrast, controlling the search of a domain independent planner requires the user to think in terms of specific "choice-points" in the planner's search strategy.

2. More importantly, search control in domain independent planners typically involves generating unpromising partial plans first and then pruning them. In contrast, our approach improves efficiency by "folding in" the control knowledge into the synthesized planner code, through incremental evaluation of pruning test. Specifically, to a first approximation, conventional domain independent planners will add the GOODNESS-TEST either to control their search at the choice points or to post-process the generated partial plans. We specify distributive laws on how the GOODNESS-TEST can be incrementally evaluated (in terms of CROSS-GOODNESS-TEST and GOODNESS-TEST) and perform context-dependent analysis on all the pieces of available knowledge to optimize the code. This is what we mean by "folding in" the control knowledge.

In Figure 11, we can see an instance of control knowledge being folded into the synthesized progression planner for the blocks world domain. Contrast it to the first synthesized planner shown in Figure 6. We notice that redundant invocation of various checks (such as GOODNESS-TEST, VISITED-STATES, etc) in the earlier planner have been simplified away. Moreover, individual checks (such as NO-MOVES-BACK and GOODNESS-TEST) have been further simplified based on the distributive laws (like CROSS-NO-MOVES-BACK and CROSS-GOODNESS-TEST) to consider just the newly added parts of the partial plan. All these considerations lead to a very small and efficient final planner.

## 5. Empirical Evaluation of Synthesized Planners

Table 1 lists several domain dependent state-space planners that we have synthesized to date. The planners are characterized by the domain for which they are developed ("BW" for blocks world, "LOG" for logistics, and "TYR" for Tyre World – all of which are benchmark domains in AI planning); the type of (state-space) refinement used ("P" for progression and "R" for regression), and the type of domain specific control knowledge used (H1 that limits useless moves, H2 which moves blocks via table, etc.). We will now report results of an empirical study conducted over these synthesized planners. The study had two aims:

- To ascertain whether the synthesized planners are able to efficiently exploit domain knowledge.

```
%% The main function PLANNER calls the auxiliary function
function PLANNER (INITIAL_STATE, GOAL_STATE, OPERATOR_SET)
  returns PLAN
  such that OUTPUT CONDITIONS are satisfied
  =
    PLANNER_AUX(INITIAL_STATE, GOAL_STATE, OPERATOR_SET,
      INITIAL_PLAN, ...)
        %% Note: other parameters are added during finite differencing

%% The function PLANNER_AUX is called by PLANNER or by PLANNER_AUX
%% Note: From the set of OUTPUT CONDITIONS given as part of
%%       the problem specification, KIDS selects all but one
%%       to be maintained as INVARIANTS. Finite differencing
%%       gives more INVARIANTS.
function PLANNER_AUX(INITIAL_STATE, GOAL_STATE, OPERATOR_SET,
          CURRENT_PLAN, ...)
  such that INVARIANTS are satisfied for the state sequence
          corresponding to the CURRENT_PLAN
  return type of PLAN
  =
    if state sequence ensures the satisfaction of GOAL_STATE
          when CURRENT_PLAN is executed from the INITIAL_STATE
       return CURRENT_PLAN
    else consider some PARTIAL_PLAN2 = append(CURRENT_PLAN, I)
          where I is index of a valid operator in OPERATOR_SET, and
          a) incremental test of INVARIANTS is true for
             state sequence corresponding to PARTIAL_PLAN2
          b) PARTIAL_PLAN2 =
                PLANNER_AUX(INITIAL_STATE, GOAL_STATE, OPERATOR_SET,
                PARTIAL_PLAN2, ...)
```

Figure 12: Pseudo-code for the state-space planners synthesized by KIDS

- To ascertain if the synthesized planners are better than traditional domain independent planners in utilizing domain control knowledge.

Our discussion is organized as follows: In Section 5.1, we describe the domains and problems we have considered in our empirical work. In Section 5.2, we evaluate the absolute performance of the synthesized planners in the various domains. In Section 5.3 we compare the performance of synthesized planners and traditional planners.

## 5.1 Domains and Problems

### 5.1.1 BLOCKS WORLD DOMAIN

A blocks world domain is an environment in which each block is placed either on a table or on top of other blocks, and the problems involve stacking them in some desired configuration. Let us focus on two states in particular: A-ON-TOP and N-ON-TOP. When there are A .. N blocks, A-ON-TOP stands for the state where block A is on top of block B, block B is on top of block C, ..., block N-1 is on top of block N and block N is on table. Similarly, N-ON-TOP stands for the state in which block A is on table, block B is on top of block A, ..., block N is on top of block N-1. Problems for the reported experiments are one of three types:

1. **Stack Inversion:** Invert from A-ON-TOP to N-ON-TOP.

2. **Stack building:** Initial state is a collection of random stacks of up to two blocks height in which the last N/2 blocks are either put on the first N/2 blocks or on the table. Goal state is A-ON-TOP or N-ON-TOP.

3. **Random blocks world problems:** A subset of random blocks world problems generated using Minton's algorithm (Minton, 1988). In a problem with N blocks, the goal state can have up to N/2 goal conditions.

| Name | Customized for the domain | Refinement | Domain Dependent Pruning Test |
|------|---------------------------|------------|-------------------------------|
| BW-P-H1 | Blocks World | Progression (FSS) | Limit useless moves (H1) |
| BW-P-H2 | Blocks World | Progression (FSS) | Move via table (H2) |
| BW-R-H1 | Blocks World | Regression (BSS) | Limit useless moves (H1) |
| BW-R-H2 | Blocks World | Regression (BSS) | Move via table (H2) |
| LOG-P-L | Logistics | Progression (FSS) | Limit inefficiency |
| LOG-R-L | Logistics | Regression (BSS) | Limit inefficiency |
| TYR-P-M | Tyre World | Progression (FSS) | Multiple control rules |
| INDEP-P | -none- | Progression (FSS) | True (every state is OK) |
| INDEP-R | -none- | Regression (BSS) | True (every state is OK) |

Table 1: Table showing the variety of planners synthesized on KIDS. The names of the planners follow the format <domain>-<refinement>-<heuristic>.

Some domain dependent pruning tests for blocks world were covered in Section 3. Specifically, we covered pruning test H1 that prevents any block from being moved in consecutive steps, and test H2 which requires that all blocks have to be moved via the table.

### 5.1.2 Logistics Domain

The logistics domain consists of a several planes and packages at different places. The goals involve transporting the planes and packages to the specified locations. We considered a type of logistics domain where there are $k$ planes, $2k$ places and $3k$ packages. There can be either 2 packages or 1 package and 1 plane at each place. The goal is to get all planes and packages to a distinguished place.

The domain dependent pruning test for the logistics transportation domain, (which we call "*Limit Inefficiency*" heuristic) consists of the following pieces of advice:

1. Planes should not make consecutive flights without loading or unloading a package

2. Packages should either be at the goal position to begin with, or may be loaded inside a plane and then flown to their goal position.

3. Once a package reaches its goal position, it should not be moved.

### 5.1.3 TYRE WORLD DOMAIN

The Tyre world (Russell & Norvig, 1995) is a benchmark domain with complex causal structure (Blum & Furst, 1995). In the "fixit" problem from this domain, a car's tyre is flat and must be replaced by a spare flat tyre (which must first be inflated). The original tyre has to be placed in the boot and all tools must be returned to the boot. The domain dependent pruning test that we devised enforces the following constraints:

1. If only one state variable changes between one state($S_1$) and the next state($S_2$), it should not change in the subsequent state($S_3$). This is because a state variable describes an attribute about an object in the domain. If the attribute has a value in state $S_2$ that will be "overwritten" in state $S_3$, this might indicate a non-minimal plan.[5]

2. Work on the status of boot last.

3. Fixing up a free hub is invalid – a wheel must be on it first.

4. If we have jacked-up the car that needs a tyre, we can not jack it down without putting the tyre on it.

5. Work on the position of the pump and wrench after all the wheels and the hub are in their final configuration.

6. Once the wheels are in their goal positions, they should not be moved.

### 5.2 Absolute Performance of the Synthesized Planners

In this section, we discuss the absolute performance of the synthesized planners in different domains. As we shall see, the synthesized planners were able to solve the benchmark problems that are known to be hard for the traditional planners. A special note is in order regarding the plots that follow. In all the plots that follow, if a curve stops mid-way in a graph, it means that the corresponding planner could not solve the given problem or all problems in the problem class (as applicable) in the stipulated time.

### 5.2.1 PLANNERS IN BLOCKS WORLD

In traditional planners, domain specific information helps the planner return a result faster and we obtained similar results with the synthesized planners. As can be seen from the left plot in Figure 13, the domain-specific blocks world heuristic H1 ("Limit useless moves") helped the progression planner (BW-P-H1) solve the stack inversion problem (invert A-ON-TOP to N-ON-TOP) for 14 blocks in under a minute, and for 22 blocks in under 30 minutes. Without such a heuristic, the progression planner (INDEP-P) could not solve even a 5 block stack inversion problem in the same time.

---

5. This is not true in all domains, since in some domains, the only way a state variable can shift its value from $v_1$ to $v_3$ is to transition through $v_2$. The heuristic however does preserve completeness in the tyre world.

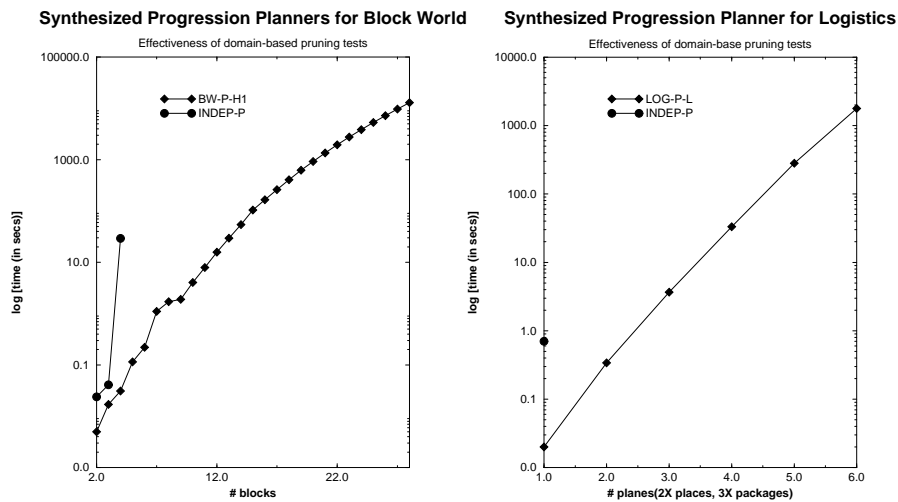

Figure 13: Plots show that domain dependent goodness tests lead to efficient planners. The problem domains are blocks world (left) and logistics transportation domain (right). Planners with domain guidance perform better than those without it.

### 5.2.2 PLANNERS IN THE LOGISTICS DOMAIN

In the logistics domain, the progression planner with the Limit Inefficiency heuristic (LOG-P-L) could solve 4-plane problems in under a minute and 6-plane problems in 30 minutes (Figure 13, right). Without such a heuristic, the progression planner (INDEP-P) could not solve even the 2 plane problem in the same time.

### 5.2.3 PLANNERS IN THE TYRE WORLD

There are 25 operators, 27 state variables and 6 control rules in our manually encoded Tyre world (Russell & Norvig, 1995) description. The fixit problem was solved in under a minute and a 31 step plan was returned.

### 5.3 Comparing Traditional and Synthesized Planners in Blocks World

Since our synthesized planners used domain specific control knowledge that is not normally used by domain independent planners, our next step involved comparing synthesized planners to domain independent planners using the same control knowledge. Our aim is to see if the synthesized planners are better able to exploit the domain knowledge than the traditional planners. We restricted this detailed comparison to the blocks world domain. Since there are a variety of traditional classical planners each of which have varying tradeoffs (c.f. (Barrett & Weld, 1994; Kambhampati et al., 1995)), we used a "league tournament" approach in our comparison. Specifically, since most popular classical planners correspond to different instantiations of UCP (Kambhampati & Srivastava, 1995), we first ran them all on our blocks world problem distribution to isolate the best traditional planners. Similar study was done to isolate the best synthesized planners for our problem distribution. At this point, the best synthesized planner is compared to the best traditional planner. In this second round of comparison, the winning traditional planner is given the same control knowledge as the synthesized planner.

We have used two of the three blocks world test suites – the random blocks world problems and the stack building problems – in the comparisons. Each problem class is defined in terms of the number of blocks and an average of 10 runs is shown in each plot. The total time allowed for a class of problems was 1000 seconds after which the planner was deemed to have failed on that problem class. All planners were run on the same problems from the problem suite.

### 5.3.1 PICKING THE BEST SYNTHESIZED PLANNER

In this section, we want to empirically determine the the most effective refinement and control knowledge (heuristics) for the blocks world problem suites. We ran six synthesized planners (BW-P-H1, BW-P-H2, INDEP-P, BW-R-H1, BW-R-H2 and INDEP-R) on the above test suite. Figure 14 shows the relative performance. We notice that planners with the pruning test H2 perform the best when compared with other planners using the same refinement.[6]

### 5.3.2 PICKING THE BEST TRADITIONAL PLANNER

In this section, we empirically search for the best UCP strategy in the blocks world. Figure 15 shows the performance of UCP instantiations with no domain dependent heuristic information. Instantiations of UCP which do only FSS, BSS or PS refinements can emulate classical forward-state space, backward-state space or plan-space planners, respectively. We call these instantiations UCP-FSS, UCP-BSS and UCP-PS. UCP-LCFR is a hybrid strategy which interleaves FSS, BSS and PS refinements depending on the lower branching factor (Kambhampati & Srivastava, 1995). In both the random blocks world problems (left) and the stacking building problems (right), the left and right plots, UCP-FSS solves all of the problems in the least time. Based on the results, we see that UCP-FSS is a good strategy for the blocks world problem distributions we used.[7]

---

6. We also notice that progression planners perform better in the left plot and regression planners perform better in the right plot. This trend can be explained easily in terms of the way the refinements operate (Kambhampati, 1997b). In the left figure, based on the nature of the goals, the branching factor for the regression planners may become enormous because it cannot detect all the conflicts among the steps that give conditions at the goal (or at steps which eventually support the goal condition). Many more operators seem to potentially give a condition than is actually the case. On the other hand, the completely specified initial state helps the progression planner decide all applicable operators from the beginning itself. Consequently, progression planner BW-P-H2 is a clear win. In the right plot, as the initial and goal states are completely specified, both regression and progression planners can detect all conflicts. As is true in realistic domains, many more operators are applicable from the initial state than are relevant to achieving the goal conditions. So, the regression planner BW-R-H2 performs better than the progression planner BW-P-H2.

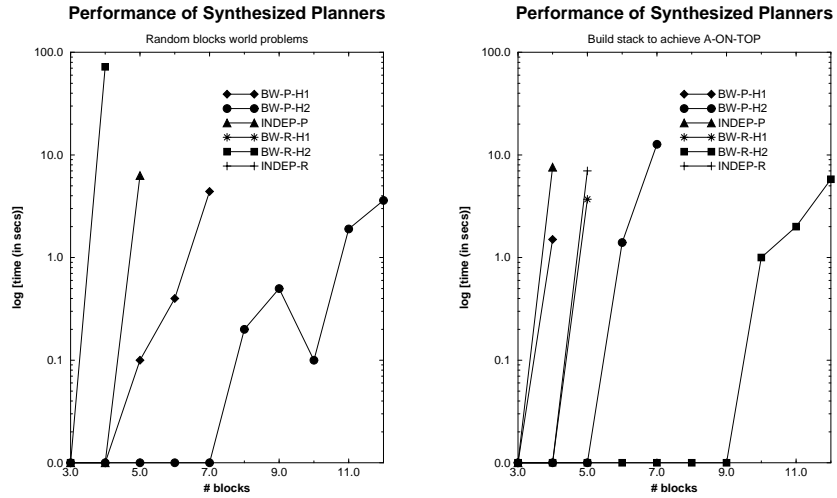

Figure 14: Performance of the synthesized progression and regression planners with H1, H2 or no domain dependent control knowledge. BW-P-H2 perform best in the left plot and BW-R-H2 performs best in the right plot. As the points are close together, it is not clear but BW-R-H1 and INDEP-R solve problems of upto 3 blocks in the left plot.

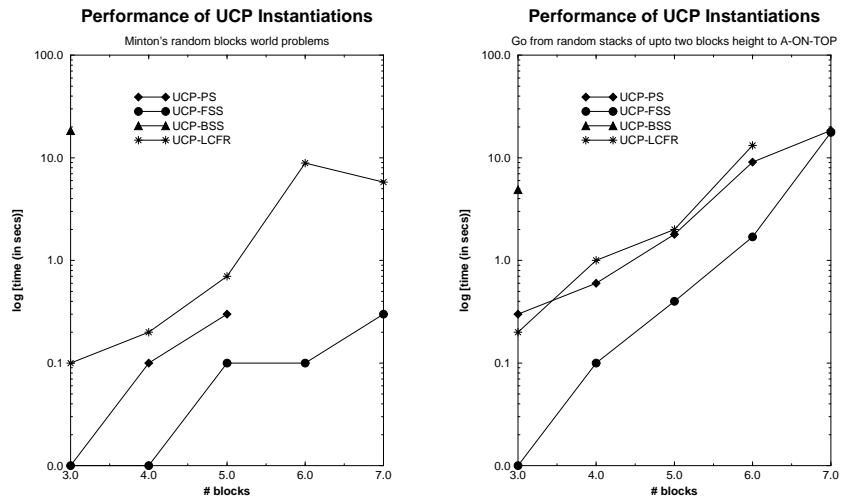

Figure 15: Performance of UCP instantiations with no domain dependent heuristic. In the left and right plots, UCP-FSS solves all of the problems in the least time. Based on the results, we see that UCP-FSS is a good strategy for blocks world. Note from the figure that UCP-BSS solves only 3 blocks problems in the given time in all the plots.

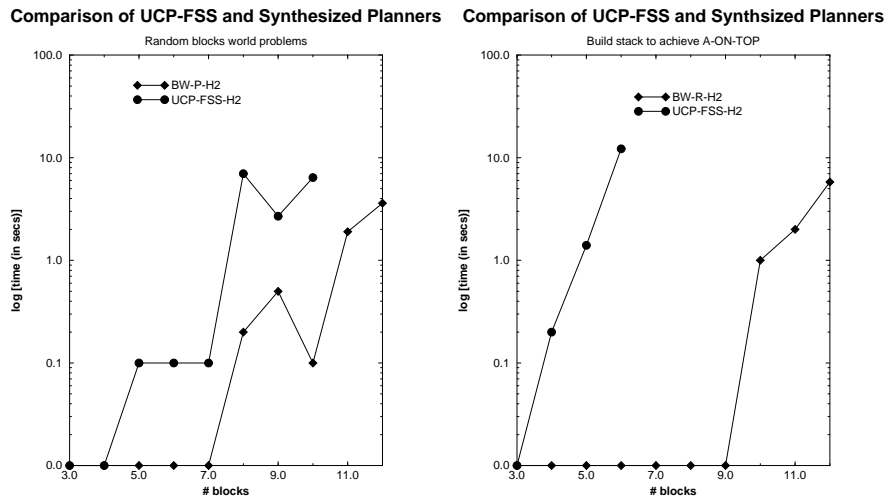

Figure 16: How the best UCP strategy for blocks world, namely UCP-FSS, performs against the best of KIDS' synthesized planners: In the left plot, UCP-FSS does better than INDEP-P i.e., without any heuristic information. But when H2 heuristic is given to both the planners, BW-P-H2 is a winner. In the right plot, BW-R-H2 outperforms UCP-FSS with H2.

### 5.3.3 COMPARING THE BEST SYNTHESIZED PLANNER AND BEST TRADITIONAL PLANNER

Finally, we pit the best UCP strategy for the blocks world, namely UCP-FSS, against the best of KIDS' synthesized planners. We chose BW-P-H2 for the random blocks world problem suite and BW-R-H2 for the stack building problem suite. Comparison is done when all planners are either given the same heuristic information (H2) or no domain dependent guidance. Figure 16 plots the results. In the left plot, BW-P-H2 is a clear winner when both planners are given the domain specific heuristic H2. In the right plot, BW-R-H2 outperforms UCP-FSS with H2. So, we see that *given the same heuristic information, the best of the planners synthesized by KIDS can outperform the best instantiation of UCP for the blocks world.*

It is interesting to note that while all synthesized planners improve drastically with domain specific knowledge, domain independent planners do not always improve in the same way. We illustrate this in Figure 17, where we compare UCP instantiations and the synthesized planners with and without control heuristics. While the synthesized planners always improve with the addition of control knowledge, the same is not true of the UCP instantiations. Specifically, in the right plot, we see that UCP-FSS-H2, which uses control knowledge H2 does worse than UCP-FSS-DOM-INDEP, which does not use any control knowledge! We speculate that this is because UCP-FSS explicitly calls a function to do domain-specific reasoning in each recursive invocation while the synthesized planners have domain control knowledge folded into the planner code (see Section 4).

---

7. Note that in the case of synthesized planners, a planner based on regression outperformed one based on progression in the stack building problems. In contrast, UCP-FSS out performs UCP-BSS for the same problem suite. The reason for this had to do with the particular implementation of UCP-BSS, which involves a costly unification step.

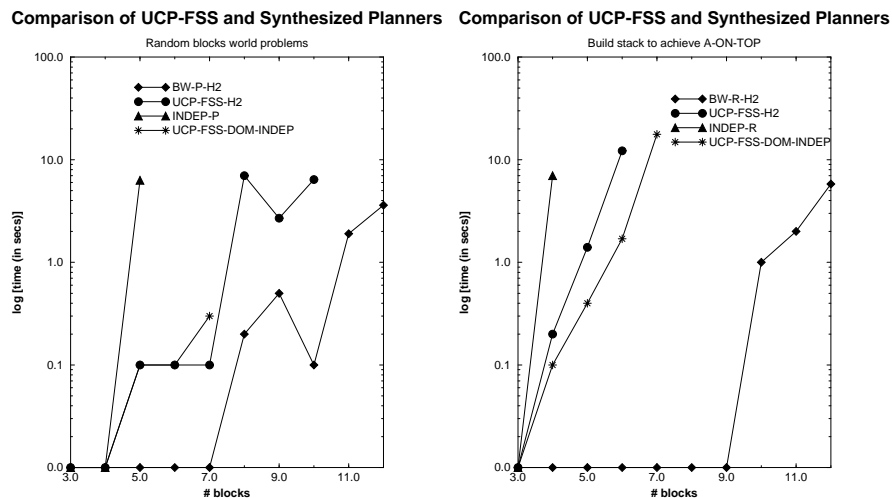

Figure 17: Effect of control knowledge on UCP planners vs. KIDS synthesized planners.

## 5.4 Summary of Results

In summary, we have demonstrated that:

- The CLAY approach to writing domain-specific planning theory is flexible. It supports mix-and-match of refinements and control knowledge in generating a variety of customized planners. As examples, we presented planners for blocks world, logistics domain and Tyre domain, using forward and backward state space refinements and different types of domain-specific control knowledge.

- The synthesized planners can exploit domain dependent control knowledge to improve performance. We showed that, in fact, they are better able to exploit domain knowledge than the traditional planners can.

## 6. Related Work

The research reported here straddles the two fields of automated software synthesis and AI planning. Although our research is the first to address the issue of planner synthesis, synthesis of other types of search engines has been addressed by Smith and his co-workers. Of particular interest to us is the work they did in developing automated scheduling software for transportation logistics problems using KIDS system (Smith & Parra, 1993; Burstein & Smith, 1996). The scheduling systems they have generated have been shown to significantly outperform general-purpose schedulers working on the same problem. Their results provided the initial impetus for our research. Although their original work used a design tactic based on global search to model scheduling, they have also since then developed design tactics based on other local search regimes. Another interesting issue brought up by their work is the importance of "constraint propagation" techniques in deriving efficient code. This has made us explore the role (or lack thereof) of constraint propagation in planning. Kambhampati and Yang (1996) describe ways in which the refinement planning framework can be extended to exploit constraint satisfaction techniques. In future, we hope to be able to synthesize planners using this more general theory of refinement planning.

Although there has not been much work on automated planner synthesis, a notable exception is the work of Gomes (1995). Gomes had synthesized a state-space problem solver for the "missionaries and cannibals" problem on KIDS, and has shown that the synthesized code outperforms general purpose problem solvers in that domain. Our framework can be seen as a generalization of the work done by Gomes. In particular, we separate planning theories from the dynamics and the control knowledge, which in principle supports generation of planning code based on a variety of refinements. We have demonstrated this by deriving both progression and regression planners for three different domains (blocks world, logistics and tyre world) and with two different bodies of control knowledge in each case. Methodologically, our work adds to Gomes' results in that we have shown that given the same control knowledge, planners generated by KIDS can outperform traditional planners using the control knowledge at their choice points. This makes for a fairer comparison between synthesized and general-purpose planners.

In some existing planners such as UCPOP (Penberthy & Weld, 1994) and PRODIGY (Fink & Veloso, 1994), customization is supported by allowing the user to specify search control rules that are checked at every choice point during search. Such control rules can be used to rank the search nodes based on some heuristic, or prune unpromising nodes. The main difference between these approaches and the synthesis approach described here is that ours supports a higher degree of integration of domain knowledge into the planner by folding it into the synthesized code for the planner. Planners using search control rules cannot do context dependent analysis or incremental application of control knowledge. In contrast, in our approach, the control information is encoded declaratively and the planning algorithm can be optimized based on *all* the knowledge that is available, including the control knowledge.

Our work on utilizing explicit control knowledge in addition to domain dynamics in deriving planning code has some parallels with the recent work by Bacchus and Kabanza (1995). They concentrate on providing a rich language in which control knowledge can be specified for a progression planner. They describe a language based on temporal logic to specify domain control rules. Rather than using this knowledge to prune bad plans after they are generated, Bacchus and Kabanza explore ways of incrementally tracking the level of satisfaction of the control axioms as the planning progresses. Our approach facilitates the same, in a refinement independent setting, by "folding in" the control knowledge into the developed planning code, with the help of KIDS framework.

There is some work in constraint satisfaction community that is directed towards producing specialized (customized) programs that is relevant to the research described here. COASTOOL (Yoshikawa, Kaneko, Nomura, & Watanabe, 1994) and ALICE systems (Lau-riere, 1978) take declarative description of CSPs and compile specialized algorithms for solving them, and MULTI-TAC (Minton, 1996) supports automatic configuration of constraint satisfaction programs. The MULTI-TAC system, in particular, provides an interesting contrast to our approach. MULTI-TAC starts with an algorithm schema, a list of high level heuristic rules for various decision points (e.g., "most-constrained-variable-first" heuristic for variable selection and "least-constraining-value-first" for value selection in CSP search), and a list of flags indicating whether certain procedures (e.g., forward checking in CSP) will or will not be used. MULTI-TAC uses the domain specification to specialize the high-level heuristics given to it. For example, in the context of a minimum maximal matching problem in graph theory, a most-constrained-variable-first heuristic may become "choose the edge with the most neighbors that have been assigned values" . A configuration is a particular subset of specialized heuristics to be used, and a particular assignment of flags. MULTI-TAC first searches through a space of "configurations" to see which configuration best fits a given problem population. Once the best configuration is found, it is then automatically compiled into efficient code by using specification refinement techniques similar to those that we described in Section 3.5.

MULTI-TAC thus presents an interesting middle-ground between search control rule specification approach used in UCPOP and PRODIGY planners, and the full integration of domain-knowledge into the synthesized code, promised by the CLAY approach. In contrast to the UCPOP and PRODIGY search control rule approach, the MULTI-TAC compilation phase can support folding-in of search control rules into the compiled code. In contrast to CLAY which advocates semi-automatic synthesis of a piece of software by manually guided optimization (through the help of user specified distributive and monotonicity laws), MULTI-TAC supports fully automating the customization of a configurable template. For the CLAY approach to be effective, we need to provide a declarative specification of the task and the domain control knowledge, as well as high level algorithm tactics. KIDS deals with instantiating the tactics into the specific problem, but the simplification needs to be guided by careful specification of distributive/monotonicity laws. In contrast, MULTI-TAC's configuration approach needs an algorithm template that is already semi-customized to the specific task, with built-in hooks for using heuristics. The heuristics themselves are specified in the form of meta-heuristic knowledge.

In MULTI-TAC, domain knowledge is used only in specializing the meta-heuristics. In theory, the CLAY approach may support a deeper integration of the domain knowledge into the synthesized code; but at the expense of a significant amount of user intervention. An interesting application of the MULTI-TAC approach in the context of planning might involve starting with a UCP planning shell (which can be configured to emulate many varieties of planners), a list of high-level heuristics for guiding the decision points in the UCP shell (e.g., refinement selection, flaw selection etc), and searching among the configurations to pick a planner for the given problem distribution.

## 7. Conclusion

In this research, we investigated the feasibility of using automated software synthesis tools to synthesize customized domain-specific planners. We described the CLAY architecture for flexibly synthesizing efficient domain dependent classical planners from a declarative theory of planning and domain theory using a software synthesis system (KIDS). Using this framework, any classical planner can be synthesized enriched with domain control knowledge. As a proof of concept, forward state-space and backward state-space planners were synthesized for the blocks-world, the logistics and the tyre world domains. We have shown that the synthesized planners can outperform general purpose planners when both are using the same amount of domain-specific control knowledge and argueed that this is due to their ability to fold-in domain specific control knowledge into the planner code. In contrast, the domain independent planners test the control knowledge for each plan being refined, and thus suffer a significant application overhead.

## 7.1 Features and Limitations

Our synthesis approach provides several interesting contrasts to main-stream AI planning work. To begin with, most AI planning work attempts to improve the efficiency of planning by concentrating on the way plans are generated. Our work differs radically in that we concentrate on how "efficient planners" are synthesized. The use of software synthesis techniques lends modularity to the planner synthesis activity. The planning theory is specified declaratively rather than in the form of an implemented program. This supports changes and extensions to the planning theory. While a planning theory is described independent of domains, control knowledge and dynamical knowledge of the domain can be specified once for each domain. By selecting different combinations of planning theory and control knowledge, we can synthesize a variety of domain-customized refinement planners.

Despite these promises, our approach does entail several overheads. Some of these overheads are related to the current state of the art in automated software synthesis while others are related to our current implementation of CLAY architecture. In what follows, we try to tease these apart.

The holy grail of automated software synthesis approaches is to free the users from low-level coding, and allow them to concentrate on declarative specification. While the KIDS system comes closest to this promise, it is still far from perfect. To start with, the user must be reasonably familiar with the software synthesis process in order to do anything substantial with KIDS. We had to go through a steep learning curve before we could understand how to structure our theories to make good use of the optimizations provided by KIDS. Writing the monotonic and distributive laws for operations such that they can help KIDS do effective code simplification is still somewhat of an art. Many times, we had to go back and rewrite the domain knowledge after KIDS was unsuccessful in using the knowledge provided to it. Advances in software synthesis technology may provide support for automatic translation for high level control knowledge into forms suitable for consumption by KIDS; but such support is not available right now.

The current cost-benefit ratios are such that we would not recommend using CLAY/KIDS approach for customizing a planner if one is interested in customizing a single planner for a single domain. The REFINE code the user writes to specify the synthesis task is typically larger than any one single synthesized planner generated by KIDS. Thus, manually customizing the planner for the domain may still be more appropriate. However, the synthesized approaches may be competitive if we are interested in being able to customize a variety of planners to a variety of domains.

In addition to the overheads entailed by KIDS, our specific implementation of planning theories, domain knowledge, etc. also lead to some inefficiencies. These latter can be eliminated by a better design of the CLAY architecture. For example, to make our work simple, we decided to go with one of the pre-existing canned design tactics provided by KIDS, and chose the global search theory over finite sequences as the candidate tactic. Because of this choice, we found state-variable representation of domains to be more suitable from an implementation point of view. Although getting state-variable representations of actions is not very hard (we wrote a couple of utility routines for converting actions in STRIPS representation into state-variable representation), specifying control knowledge in terms of this representation turned out to be less than natural, especially in larger domains like Russel's tyre world.

Our choice of in-built design tactics also limited the types of domain knowledge we could specify. Most of the control knowledge had to be in the form of node-rejection rules. General global search also allows node-preference knowledge as well as knowledge regarding effective ways of shrinking the set of potential solutions, without splitting the set (by eliminating non-solutions).

We could eliminate the awkwardness of state-variable representations as well as exploit more types of domain knowledge by designing global search tactics specially suited to planning-specific data structures. Although eventually the KIDS system may support a larger variety of design tactics, customizing design tactics to task classes is very much in line with the current practice in automated software synthesis (Gomes, Smith, & Westfold, 1996).

## 7.2 Future Directions

The work presented here can be seen as the beginning of a fairly open-ended research program that complements, rather than competes with, the research into efficient planning algorithms. Ideally, we would like to support the synthesis of customized planners based on the full gamut of planning technologies including partial order and task-reduction planning. These latter are already subsumed by the refinement planning framework developed in (Kambhampati & Srivastava, 1996) and supporting their synthesis is mainly a matter of supporting a more flexible partial plan representation in KIDS (representing plans as sequences over actions has sufficed until now, as we were only addressing the synthesis of state-space planners). We are currently in the process of doing this (Srivastava, Kambhampati, & Mali, 1997).

More generally, any time we get insights into the internal workings of a family of planning algorithms, we would like to translate those insights into declarative specifications for KIDS and support synthesis of more efficient customized domain code. An example of this is the recent research on plan synthesis approaches based on constraint satisfaction. In fact, domain independent planners such as Graphplan (Blum & Furst, 1995) can solve our test suites in equal or better time compared to the synthesized planners. We have taken some preliminary steps towards integrating these approaches into the refinement planning framework by using the notion of disjunctive refinement planning (see Kambhampati & Yang, 1996; Kambhampati, 1997b; Kambhampati, Parker, & Lambrecht, 1997; Kambhampati, 1997a). In future, as this work matures, we intend to explore synthesis of planners using the theories of disjunctive plan refinement.

## Acknowledgements

We are grateful to Doug Smith for his help with KIDS and numerous discussions on tactics, global search and distributive laws. We would also like to thank Carla Gomes for discussing her work with us and making useful suggestions on how to write theories; and Nort Fowler for his encouragement on this line of research. Thanks are also due to Steve Minton and the anonymous referees of *JAIR* for their many helpful comments toward improving the presentation of this paper. This research is supported by a DARPA Planning Initiative Phase 3 grant F30602-95-C-0247 (through a subcontract from Kestrel to Arizona State University).

## Appendix A. Sample Code Referred in the Text

```
%% Get state sequence from partial plan
function VISITED-STATES
  (PLAN: seq(integer), IS: seq(integer),
   GS: seq(integer),
   OPERS: seq(tuple(seq(integer), seq(integer))))
  : seq(seq(integer))
  = if empty(PLAN) then [IS]
      else
      if ~empty(NEXT-STATE(IS, first(PLAN), OPERS)) then
        prepend(VISITED-STATES(rest(PLAN),
                  NEXT-STATE(IS, first(PLAN), OPERS),
                  GS, OPERS), IS)

%% Used by VISITED-STATES to get next state
%% after applying an operator
function NEXT-STATE
  (STATE: seq(integer), OPER: integer,
   OPERATORS: seq(tuple(seq(integer), seq(integer))))
  : seq(integer)
  = if fa (I: integer)(I in [1 .. size(STATE)]
        => OPERATORS(OPER).1(I) = 0 or
           OPERATORS(OPER).1(I) = STATE(I)) then
      image(lambda(x)
              if OPERATORS(OPER).2(x) = 0 then STATE(x)
              else OPERATORS(OPER).2(x),
              [1 .. size(STATE)])

%% Domain independent pruning test for FSS refinement
function NO-MOVES-BACK (VIS-STATES: seq(seq(integer)),
  IS: seq(integer), GS: seq(integer))
  : boolean
  = fa (I: integer, J: integer)
       (I in [1 .. size(VIS-STATES)]
         & J in [1 .. size(VIS-STATES)]
         & I < J
         => not (fa (INDEX: integer)
                  (INDEX in [1 .. size(first(VIS-STATES))]
                    => VIS-STATES(J)(INDEX) =
                                VIS-STATES(I)(INDEX)
                    or VIS-STATES(I)(INDEX) = 0)))
```

```
function CROSS-NO-MOVES-BACK
  (R: seq(seq(integer)), S: seq(seq(integer)), IS: seq(integer),
  GS: seq(integer)): boolean
 = fa (I: integer, J: integer)
     (I in [1 .. size(R)] & J in [1 .. size(S)]
        => not (fa (INDEX: integer)
                 (INDEX in [1 .. size(first(R))]
                   => S(J)(INDEX) = R(I)(INDEX)
                      or R(I)(INDEX) = 0)))

%% Domain independent goal test for FSS refinement
function GOAL-TEST
  (VIS-STATES: seq(seq(integer)), INIT: seq(integer),
  GOAL: seq(integer)): boolean
 = fa (I: integer)
    (I in [1 .. size(GOAL)]
       => GOAL(I) = 0 or
          last(VIS-STATES)(I) = GOAL(I))

%% Domain dependent pruning test
%% Heuristic H1: Limit useless move for blocks world
function NO-REDUNDANCY (S: seq(seq(integer)))
: boolean
= fa (I: integer, INDEX: integer)
   (I in [1 .. size(S) - 2]
       & INDEX in [1 .. size(first(S))]
       & INDEX mod 2 = 1
       and S(I)(INDEX) ~= S(I + 1)(INDEX)
       => S(I + 2)(INDEX) = S(I + 1)(INDEX))

%% Domain dependent pruning test
%% Heuristic H2: Move via table for blocks world
function NO-REDUNDANCY (S: seq(seq(integer)),
                        INIT: seq(integer),
                        GOAL: seq(integer))
 : boolean
 = fa (I: integer, INDEX: integer)
    (I in [1 .. size(S) - 1]
      & INDEX in [1 .. size(first(S))]
      & INDEX mod 2 = 1
      & S(I)(INDEX) ~= S(I + 1)(INDEX)
     =>((%% position in state I is same as
        %% initial state
        S(I)(INDEX) = INIT(INDEX)
        and
        S(I + 1)(INDEX) =
         %% position in state I+1 is on table
         (1 + real-to-nearest-integer
                   (size(first(S)) / 2)))
      or
        (S(I)(INDEX) =
          %% position in state I is on table
          (1 + real-to-nearest-integer
                    (size(first(S))/2))
        and
        %% position in state I+1 is same as
        %% goal state
        S(I + 1)(INDEX) = GOAL(INDEX))))
```